\documentclass[11pt]{article}
\usepackage{gscl2015}
\usepackage{times}
\usepackage{url}
\usepackage{latexsym}
\usepackage{graphicx}
\usepackage[english]{babel}

\title{Sentiment Uncertainty and Spam in Twitter Streams and Its Implications for General Purpose Realtime Sentiment Analysis}

\author{Nils Haldenwang\\
  University of Osnabr\"uck, Germany \\
  {\tt nils.haldenwang@uos.de} \\\And
  Oliver Vornberger\\
  University of Osnabr\"uck, Germany \\
  {\tt oliver@uos.de} \\}

\date{}

\begin{document}
\maketitle
\begin{abstract}

  State of the art benchmarks for Twitter Sentiment Analysis do not consider
  the fact that for more than half of the tweets from the public stream a
  distinct sentiment cannot be chosen. This paper provides a new perspective on
  Twitter Sentiment Analysis by highlighting the necessity of explicitly
  incorporating uncertainty. Moreover, a dataset of high quality to evaluate
  solutions for this new problem is introduced and made publicly
  available\footnote{\url{http://project2.cs.uos.de/TweeDOS}}.

\end{abstract}

\section{Introduction}
\label{sec:introduction}

As a field of research Twitter Sentiment Analysis  has gained much attention
recently. For a multitude of applications such as sales prediction
\cite{asur2010predicting}, stock market prediction \cite{bollen2011twitter} or
political debate analysis \cite{diakopoulos2010characterizing} it has been
shown to generate practical value.  Twitter Sentiment Analysis denotes
the task of assigning a given tweet a sentiment label of either
\textit{positive} or \textit{negative} and is an integral part of many
practical applications.  Few methods consider \textit{neutral} as a third
class. However, defining a neutral class is a hard task.
\newcite{pak2010twitter} for example label tweets of popular news sites as
neutral. This assumption is not always true. The headline ``Multiple children
were killed in the attack.'' would be labeled as \textit{negative} by most
human labelers.  Thus, we propose an alternative approach to this problem.
Its basic idea is the explicit incorporation of sentiment uncertainty.

\section{The State of the Art and Its Shortcomings}
\label{sec:stateOfTheArt}

SemEval-2014 Task 9 \cite{rosenthal2014SemEval} provides a widely used state of
the art benchmark for Twitter Sentiment Analysis and compares the performance
of many current approaches.  From a dataset collected from January 2012
to January 2013, popular topics have been extracted through identification of
frequently mentioned named entities. Only tweets scoring above a certain
polarity threshold determined by a sentiment lexicon were considered to ensure
the inclusion of a sentiment. The labels included in the dataset are
\textit{positive}, \textit{negative} and \textit{neutral}, determined by a
majority vote of five labelers who were told to vote for the sentiment they
perceive as strongest, when in doubt.  This assigns tweets to the classes
\textit{positive} and \textit{negative} which do not carry a distinct
sentiment.  Methods performing well on this dataset are shown to be able to
distinguish between positive and negative sentiment under the assumption that
all tweets can be assigned one of these labels. Moreover, all test tweets
include popular named entities of the time.  As the authors themselves noted:
The dataset is biased.  Moreover, the majority vote along with the treatment of
ambiguity adds noise to the dataset.  While providing a dataset of high quality
for the desired purpose, the general composition of the public Twitter stream
is not represented by the dataset.  Hence, only part of the problems arising in
practical analysis of the live stream are addressed with the related research.

\section{A General Purpose Dataset}
\label{sec:CreatingAGeneralPurposeDataset}

When analysing the Twitter stream we are interested in the ``Electronic Word of
Mouth''\cite{jansen2009twitter}, i.e. the personal opinions of private Twitter
users. While labeling tweets, we noticed that a relatively high percentage of
tweets are spam, advertising or marketing messages which we are not interested
in. Those tweets shall be labeled \textit{spam}. Moreover, it became obvious that
for the remaining tweets only a small fraction can be distinctly labeled as
\textit{positive} or \textit{negative}. The remaining tweets may still
include polarity and can often not be labeled \textit{neutral} while being
neither \textit{positive} nor \textit{negative}. Hence, we propose the new
category \textit{uncertain}. Tweets labeled as \textit{neutral} can be assigned
to the class \textit{uncertain} too, as they provide no additional information
for sentiment analysis and can be treated in the same way as tweets of
\textit{uncertain} sentiment. This approach reduces the noise for the sentiment
bearing classes which is a desirable feature if political or business decisions
are supposed to be supported by the analysis results.

To acquire a representative view on the label composition of the public Twitter
stream, we randomly sampled our dataset from a collection of about 43 million
tweets with their creation dates ranging from June 2012 to August 2013 to
minimize topical bias.  Each tweet was labeled by two human labelers who had to
assign it one of the labels \textit{positive}, \textit{negative},
\textit{uncertain} or \textit{spam}. In total 14506 tweets have been labeled by
27 labelers. The labelers consisted of master's students from the
University of Osnabr\"uck, Germany and researchers from our group.

\begin{figure}[htb]
  \centering
  \includegraphics[width=0.49\textwidth]{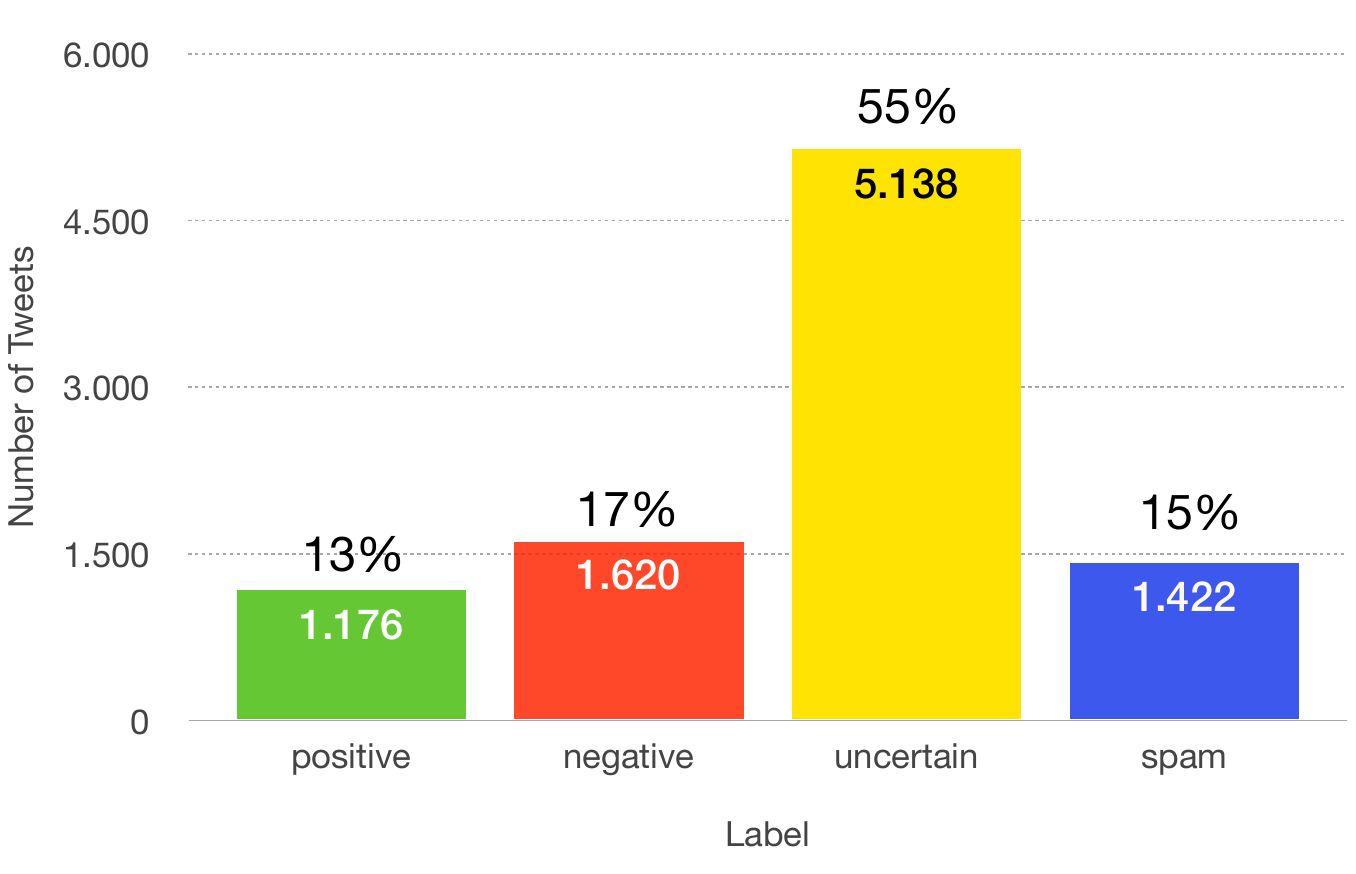}
  \caption{Distribution of labels for tweets which both labelers agreed upon.}
  \label{fig:LabelDistribution}
\end{figure}

The distribution of labels is shown in figure \ref{fig:LabelDistribution}.
There is a total of 9356 (64.5\% of total tweets labeled) tweets to which both
human labelers assigned the same label. Of these tweets 15\% are \textit{spam}
and 55\% are labeled \textit{uncertain}.  A definite sentiment label could only
be assigned to 30\% of tweets with 13\% being positive and 17\% being negative.

These results provide evidence for our claim that one has to deal with
uncertainty in sentiment analysis when working with the public Twitter
stream.

To assess the inter annotator agreement we computed Fleiss' Kappa
\cite{fleiss1971measuring} resulting in a value of $\kappa = 0.45$ which can be
interpreted as \textit{moderate agreement} \cite{landis1977measurement}. At
first sight this value seems to be rather low but when considering the
disagreement matrix shown in table 1 the claim of the
necessity to deal with uncertainty is further strengthened.

\begin{table}[ht]
\centering
\begin{tabular}{c | c  c  c  c}
           & positive & negative & uncert. & spam \\
  \hline
positive   & 1176 & 106 & 1666 & 143 \\
negative   &  & 1620 & 2263 & 58 \\
uncert.    &  &   & 5138 & 914 \\
spam       &  &   &  & 1422
\label{table:disagreement}
\end{tabular}
\caption{Disagreement matrix showing the absolute number of label combinations.}
\end{table}

Labelers seem to have a very good understanding of what distinguishes the
classes \textit{positive} and \textit{negative}, only 106 tweets have been
assigned both these labels. The disagreement for \textit{positive/spam} and
\textit{negative/spam} is of similar or even smaller magnitude. Looking at
these tweets we noticed that the disagreement is mainly related to
misunderstanding of the labeling instructions or probably accidentally clicking
the wrong label. Hence, these tweets should be omitted from the test set when
evaluating methods for reliable Twitter Sentiment Analysis.

However, the disagreement between \textit{positive/negative} and
\textit{uncertain} is relatively large.  These tweets make up about 76\% of the
tweets to which the two labelers assigned different labels.  This indicates
that in many cases not even two humans can agree upon whether a tweet contains
a distinct sentiment or should be labeled \textit{uncertain}. Systems aiming to
perform reliable sentiment analysis of the public Twitter stream should be able
to deal with these tweets.  While not strictly belonging to the category
\textit{uncertain} they should still be labeled as such or at least not be
considered for sentiment analysis.  Another possible approach can be to
interpret them as \textit{rather positive} or \textit{rather negative},
depending on the amount of reliability the respective application requires.

Moderate disagreement (914 tweets) can be noted for the classes
\textit{uncertain} and \textit{spam}. Since these tweets may still contain
useful information in the sense of answering the question ``What do people talk
about?'' they probably should not be considered spam. However, they also should
not be assigned a sentiment. A system labeling these as \textit{uncertain} will
still produce reliable results with regard to sentiment analysis.

As a first approach one can make use of just the tweets with two identical
labels to asses methods for reliable sentiment analysis of the public Twitter
stream. However, it should be considered that in practice the tweets upon which
the labelers disagreed can also appear in the stream and have to be handled to
provide reliable sentiment results. To enable researchers to develop systems
which meet all the aforementioned requirements the complete dataset including
the tweets disagreed upon is publicly available.

\section{Conclusion and Outlook}
\label{sec:ConclusionAndOutlook}

When performing analysis on the public live stream of Twitter with regard to
sentiment, it needs to be considered that more than half of the tweets cannot
be assigned a distinct sentiment. These tweets have to be filtered or
explicitly dealt with before sentiment analysis takes place. Moreover, one has
to deal with spam tweets.  Spam adds unwanted noise by polluting topics with
artificially injected tweets.  Most of the work on spam detection on Twitter
focusses on catching the users generating the spam by looking at the accounts'
behaviour over time \cite{grier2010spam,lin2013study}. When performing realtime
analysis, a given tweet has to be determined to be spam or no spam by looking
at its content and meta data only as there is no time to examine the author's
account in detail.  New methods have to be developed which are able to deal
with sentiment uncertainty and spam if reliable representations of the public
opinion are to be acquired from the Twitter stream. The dataset presented in
this paper can be used to develop and evaluate methods for reliable Twitter
Sentiment Analysis.

\bibliographystyle{gscl}
\bibliography{literatur}

\end{document}